%% file: root.tex
\documentclass[letterpaper, 10 pt, journal, twoside]{IEEEtran}

\IEEEoverridecommandlockouts                              

\input{preamble.tex}

\title{
Distributed Motion Planning with Safety Guarantees for Self-Reconfiguring Robotic Boats
}

\author{Alejandro Gonzalez-Garcia, Wei Wang, Wei Xiao, Wilm Decr\'e, Jan Swevers, Carlo Ratti and Daniela Rus%
\thanks{This work was supported by a grant from the Amsterdam Institute for Advanced Metropolitan Solutions (AMS) in Netherlands, and by the MIT-Belgium - KU Leuven Seed Fund from MIT International Science and Technology Initiatives (MISTI). A. Gonzalez-Garcia was supported by the Flanders Make SBO project ARENA (Agile \& Reliable Navigation). }
\thanks{A. Gonzalez-Garcia, W. Decr\'e and J. Swevers are with MECO Research Team, Department of Mechanical Engineering, KU Leuven, Belgium and with Flanders Make@KU Leuven, Belgium. {\tt\small \{alex.gonzalezgarcia, wilm.decre, jan.swevers\}@kuleuven.be}}
\thanks{W. Wang is with the Marine Robotics Lab, Department of Mechanical Engineering, College of Engineering, University of Wisconsin-Madison, Madison, WI 53706 USA. {\tt\small wwang745@wisc.edu}}%
\thanks{W. Xiao is with School of Electrical and Electronic Engineering, Nanyang Technological University, and with M3S, SMART, Singapore. {\tt\small wei.xiao@ntu.edu.sg, weixy@mit.edu}}%
\thanks{C. Ratti is with the SENSEable City Laboratory, Massachusetts Institute of Technology, Cambridge, MA 02139 USA. {\tt\small ratti@mit.edu}}
\thanks{D. Rus is with the Computer Science and Artificial Intelligence Lab (CSAIL),  Massachusetts Institute of Technology, Cambridge, MA 02139 USA. {\tt\small rus@mit.edu}}%
}

\begin{document}

\maketitle
\thispagestyle{empty}
\pagestyle{empty}

\begin{abstract}

Aquatic self-reconfigurable robots must assemble into desired shapes while ensuring safe interactions among multiple agents. This paper proposes a hybrid framework that combines distributed Model Predictive Control (MPC) with Control Barrier Functions (CBFs) for multi-agent shape formation and reconfiguration. Given a desired shape and target assignment, a distributed MPC scheme, solved via the Alternating Direction Method of Multipliers (ADMM), computes coordinated trajectories through local optimization and information exchange. To ensure safety in real time, distributed CBF-based filters are applied to enforce inter-agent collision avoidance. The proposed approach leverages the predictive capabilities of MPC to mitigate local minima, while CBFs provide formal safety guarantees despite the nonconvexity of the underlying optimization problem. Simulation results with up to 25 agents and experimental validation with four physical robots demonstrate the effectiveness and scalability of the framework.
\end{abstract}

\section{Introduction}
\input{Sections/introduction}

\section{Preliminaries}
\label{sec:preliminaries}
\input{Sections/preliminaries}

\section{Methodology}
\label{sec:methodology}
\input{Sections/methodology}

\section{Numerical Simulations}
\input{Sections/scalability}

\section{Experimental Validation}
\input{Sections/experiments}

\section{Conclusion \& Future Work}

This paper presented a hybrid control framework for the self-assembly of modular robotic boats. The approach combines a centralized initial target assignment with a distributed control layer, fusing the predictive planning of ADMM with the safety guarantees of control barrier functions. Simulation and physical hardware experiments demonstrated that this framework successfully achieves collision avoidance with deadlock resolution, enabling safe execution at 5 Hz on onboard microcomputers. Future work will focus on validating the system with larger physical swarms and on three key extensions: collective transport of assembled structures, navigation in constrained aquatic environments, and robust operation under environmental disturbances.


\bibliographystyle{IEEEtran}
\bibliography{References}

\end{document}

%% file: preamble.tex
\usepackage{graphicx} 
\usepackage{times} 
\usepackage{amsmath} 
\usepackage{amssymb}  
\usepackage{cite}
\usepackage{algorithm}
\usepackage[shortlabels]{enumitem}
\usepackage{bm}
\usepackage{algorithmic}
\usepackage{booktabs}
\usepackage{multirow}

\newtheorem{definition}{\textbf{Definition}}

\newtheorem{assumption}{\rm\textbf{Assumption}}
\newtheorem{remark}{\rm\textbf{Remark}}

\usepackage{tikz}
\usetikzlibrary{shapes.geometric, arrows.meta, positioning, fit, calc, backgrounds}

%% file: Sections/introduction.tex
\label{sec:introduction}

Modular self-reconfigurable robotic systems (MSRRs) consist of multiple robotic units that can physically connect and rearrange to form larger structures \cite{Yim2007,Liu2019,Zhao2022,Yang2024,Liang2025}. This paradigm is particularly relevant in aquatic environments, where deploying and maintaining fixed infrastructure is costly and often impractical. In such settings, teams of floating robotic modules can assemble into functional configurations, supporting applications such as environmental monitoring and offshore operations by building modular platforms or temporary infrastructure \cite{SoftRafts,Zhang2025,floatform_natcomms,OHara2014,Paulos2015}. The ability to disassemble and reconfigure allows adaptation to changing tasks and environmental conditions. However, achieving reliable self-assembly remains challenging, as it requires coordinating the motion of many agents while avoiding collisions and creating physical connections.

Several existing aquatic MSRR systems leverage centralized methods to create or actuate aquatic structures, such as \cite{Paulos2015,Gheneti2019,Knizhnik2022}. In \cite{Paulos2015}, automated assembly was achieved by creating motion plans to connect vessels in sequence, demonstrating structures with up to 33 real modules. Optimization-based methods were applied in \cite{Gheneti2019}, focusing on the shapeshifting reconfiguration task. Collective locomotion using centralized control was addressed in \cite{Knizhnik2022}, allowing underactuated boats to navigate while connected in lattice structures. These approaches rely on centralized coordination, which scales poorly with the number of agents and introduces single points of failure. In contrast, \cite{Wei2020} proposed a distributed approach for collective motion, where a leader vessel guides the group while follower vessels implement consensus-based controllers using only local sensing. This suggests that distributed coordination enables scalable, real-time operation using only local information.

Recent works have explored hybrid coordination architectures that combine centralized task allocation with distributed execution to improve scalability~\cite{Zhang2025,floatform_natcomms}. In \cite{Zhang2025}, a structured planning approach is used to enable parallel assembly by decomposing the global task into coordinated subproblems, allowing multiple modules to move and assemble simultaneously. In \cite{floatform_natcomms}, a hybrid framework is proposed in which a centralized assignment is combined with distributed potential-field-based controllers, enabling fully parallel motion using local interactions between modules. However, the central target assignment was essential as potential fields alone were insufficient the perfect square-lattice structures needed for inter-agent docking. While these approaches improve scalability and enable parallel assembly, they rely on heuristics that may suffer from local minima and do not provide formal safety guarantees.

Optimization-based methods offer an alternative to move beyond heuristic-based formulation. Model Predictive Control (MPC) provides a natural framework for multi-agent coordination, as it explicitly accounts for system dynamics and constraints while enabling anticipatory behavior through finite-horizon prediction \cite{MAYNE2000789}. In multi-agent settings, MPC allows agents to coordinate their motion while satisfying coupling constraints such as collision avoidance and formation objectives. However, centralized MPC formulations require solving a large-scale optimization problem with global information, which becomes computationally prohibitive as the number of agents increases \cite{Shorinwa2024tutorial}. To address this limitation, distributed MPC approaches decompose the global problem into local subproblems that are coordinated through shared variables or constraints~\cite{Shorinwa2024tutorial}. Among these methods, the Alternating Direction Method of Multipliers (ADMM)~\cite{boyd2011admm} has been widely adopted due to its ability to handle coupled optimization problems while enabling parallel computation~\cite{ruben2017decentralised}. Despite these advantages, the resulting multi-agent optimization problem is inherently nonconvex due to the collision avoidance constraints. In this setting, ADMM-based distributed MPC does not guarantee convergence to a feasible solution, meaning that collision constraint satisfaction cannot be ensured. This limitation is particularly critical in safety-critical multi-agent systems, where violations of collision constraints must be prevented.

Control Barrier Functions (CBFs) \cite{ames2019cbf, xiao2021hocbf} provide formal safety guarantees through forward invariance of safe sets. In multi-agent systems, CBFs have been used to enforce pairwise collision avoidance constraints in a decentralized manner by filtering control inputs \cite{multiCBF}. However, when used in isolation, CBF-based controllers are typically myopic and may lead to deadlocks in dense scenarios, as they do not account for task-level objectives or long-horizon coordination.

Motivated by these complementary properties, we propose to combine distributed MPC and CBF-based safety filtering within a unified framework. In the proposed approach, ADMM-based distributed MPC is used to compute coordinated trajectories by allowing each agent to solve a local optimization problem while exchanging information with neighboring agents. These trajectories provide task-driven, predictive guidance that mitigates deadlocks arising from purely reactive control. In parallel, a distributed CBF-based safety filter is applied at the control level to ensure pairwise collision avoidance in real time, providing safety guarantees that cannot be ensured by the nonconvex distributed optimization alone. The main contributions of this work are summarized as follows:
\begin{itemize}
    \item A distributed ADMM-CBF framework for multi-agent shape formation and reconfiguration of aquatic modular robots, combining distributed trajectory optimization with safety guarantees.
    \item An extensive simulation study validating the real-time feasibility of the framework for swarms of up to $N=25$ agents, accompanied by a scalability analysis.
    \item Experimental validation with a team of four physical robots performing multi-shape reconfiguration tasks.
\end{itemize}

%% file: Sections/preliminaries.tex

In this section, the robotic platform is described, the kinematic model used for planning is introduced, the relevant control barrier function theory is summarized, and the problem is formulated.

\subsection{Modular Robotic Boats}
\label{sec:floatform}

The system is a group of miniature modular robotic boats (Fig.~\ref{fig:experiment}) designed for self-assembly and self-reconfiguration on the water surface. Each module has a square footprint of $L = 0.21$\,m, weighs approximately $0.8$\,kg, and is equipped with four miniature thrusters in a diamond configuration, enabling omnidirectional planar motion. An origami-inspired magnetic latching mechanism allows modules to physically connect on all four sides, forming rigid square-lattice assemblies. Onboard sensing consists of two ultrasonic beacons and an inertial measurement unit (IMU), processed through a Kalman filter to estimate position, heading, and velocity. A low-level feedback controller based on proportional-integral-derivative (PID) control and feedback linearization regulates surge, sway, and yaw, tracking velocity references generated by higher-level algorithms~\cite{floatform_natcomms}. Each module runs identical software on a Raspberry Pi single-board computer, using ROS~\cite{Quigley09}, and inter-agent communication is achieved through a Wi-Fi router.

\subsection{Kinematic Model}
\label{sec:kinematic_model}

Consider a group of $N$ modules operating on a planar water surface. Let $\bm{p}_i = [x_i,\, y_i]^\top \in \mathbb{R}^2$ denote the position of the $i$-th module in the inertial reference frame, and $\bm{u}_i = [u_{x,i},\, u_{y,i}]^\top \in \mathbb{R}^2$ denote its velocity input. Since the low-level controller closes the velocity loop at a significantly higher frequency than the planning layer, the closed-loop dynamics of each module are well approximated by a single-integrator kinematic model:
\begin{equation}
\label{eq:single_integrator}
    \dot{\bm{p}}_i = \bm{u}_i, \quad i \in \mathcal{N} := \{1, \ldots, N\},
\end{equation}
subject to velocity bounds $\|\bm{u}_i\|_\infty \leq u_{\max}$.

\begin{remark}
The boats are governed by nonlinear hydrodynamics~\cite{fossen_handbook}. However, the feedback-linearized velocity controller effectively compensates for the nonlinear dynamics, allowing~\eqref{eq:single_integrator} to serve as a valid planning model. The low-level controller handles disturbances from self-motion and neighboring modules at a higher frequency. We choose this hierarchical structure given the absence of  accurate dynamic model parameters.
\end{remark}

\subsection{Distributed Safety via Control Barrier Functions}
\label{sec:cbf_theory}

Control barrier functions provide a framework to enforce safety constraints on control systems through forward invariance of a safe set~\cite{ames2019cbf}. Consider a control-affine system $\dot{\bm{x}} = f(\bm{x}) + g(\bm{x})\bm{u}$, and a continuously differentiable function $h : \mathbb{R}^n \to \mathbb{R}$ defining the safe set $\mathcal{C} = \{\bm{x} \in \mathbb{R}^n : h(\bm{x}) \geq 0\}$.

\begin{definition}[Control Barrier Function~\cite{ames2019cbf}]
\label{def:cbf}
A continuously differentiable function $h : \mathbb{R}^n \to \mathbb{R}$ is a \emph{control barrier function} for the system $\dot{\bm{x}} = f(\bm{x}) + g(\bm{x})\bm{u}$ on the set $\mathcal{C}$ if there exists an extended class-$\mathcal{K}_\infty$ function $\alpha$ such that:
\begin{equation}
\label{eq:cbf_condition}
    \sup_{\bm{u} \in \mathcal{U}} \left[ L_f h(\bm{x}) + L_g h(\bm{x}) \bm{u} + \alpha(h(\bm{x})) \right] \geq 0, \quad \forall \bm{x} \in \mathcal{C},
\end{equation}
where $L_f h$ and $L_g h$ denote the Lie derivatives of $h$ along $f$ and $g$, respectively.
\end{definition}

Any Lipschitz continuous controller satisfying~\eqref{eq:cbf_condition} renders $\mathcal{C}$ forward invariant, ensuring that $h(\bm{x}(t)) \geq 0$ for all $t \geq 0$ provided $h(\bm{x}(0)) \geq 0$~\cite{ames2019cbf}. For the single-integrator model~\eqref{eq:single_integrator}, the inter-agent safety constraint between modules $i$ and $j$ is naturally encoded by the barrier function:
\begin{equation}
\label{eq:h_ij}
    h_{ij}(\bm{p}_i, \bm{p}_j) = \|\bm{p}_i - \bm{p}_j\|^2 - d^2,
\end{equation}
where $d > 0$ is the minimum allowed distance between module centers. Since the continuous-time derivative of the barrier function depends on the velocities of both agents, the global safety constraint taking $\alpha(h) = \gamma h$ with $\gamma > 0$ becomes:
\begin{equation}
\label{eq:cbf_global}
    2(\bm{p}_i - \bm{p}_j)^\top (\bm{u}_i - \bm{u}_j) + \gamma \left( \|\bm{p}_i - \bm{p}_j\|^2 - d^2 \right) \geq 0.
\end{equation}

In a decentralized setting, agent $i$ must compute its control input without instantaneous knowledge of agent $j$'s commanded velocity $\bm{u}_j$. Following the decentralized barrier certificate formulation proposed by~\cite{multiCBF}, this joint constraint can be formally decoupled by partitioning the responsibility for collision avoidance. By introducing a sharing parameter $\beta \in (0,1)$, the global constraint is split such that agent $i$ strictly enforces:
\begin{equation}
\label{eq:cbf_constraint_ij}
    2(\bm{p}_i - \bm{p}_j)^\top \bm{u}_i + \beta\gamma \left( \|\bm{p}_i - \bm{p}_j\|^2 - d^2 \right) \geq 0.
\end{equation}
Symmetrically, agent $j$ enforces its respective constraint using $(1-\beta)\gamma$. For homogeneous swarms, $\beta$ is typically set to $0.5$ to equally distribute the avoidance effort.

\begin{remark}
Since the sum of the decoupled constraints~\eqref{eq:cbf_constraint_ij} for agents $i$ and $j$ recovers the global constraint~\eqref{eq:cbf_global}, the decentralized formulation preserves forward invariance using only the current positions of neighboring modules \cite{multiCBF}.
\end{remark}

\begin{remark}
The distributed safety formulation above is derived for single-integrator dynamics, but analogous constructions exist for other control-affine models. For systems with higher relative degree, such as double-integrator or full dynamic models, higher-order CBFs can be employed to enforce the same pairwise safety guarantees~\cite{xiao2021hocbf}.
\end{remark}

\subsection{Problem Formulation}
\label{sec:problem}

The self-reconfiguration problem consists of steering $N$ modules from arbitrary initial positions to a target shape, defined as a set of lattice positions $\mathcal{G} = \{\bm{g}_1, \ldots, \bm{g}_N\} \subset \mathbb{R}^2$ with uniform spacing matching the module side length $L$. The reconfiguration task proceeds as a sequence of target shapes $\mathcal{G}^{(1)}, \mathcal{G}^{(2)}, \ldots$, where each transition requires disassembly from the current shape, safe navigation to the new configuration, and reassembly via latching.

Formally, the objectives are:
\begin{enumerate}
    \item[\textbf{O1.}] \emph{Shape formation:} Each module $i$ reaches an assigned target position $\bm{g}_{\sigma(i)}$, where $\sigma : \mathcal{N} \to \mathcal{N}$ is a target assignment;
    \item[\textbf{O2.}] \emph{Inter-agent safety:} For all $i \neq j$ and all $t \geq 0$:
    \begin{equation}
    \label{eq:safety_constraint}
        \|\bm{p}_i(t) - \bm{p}_j(t)\| \geq d(t),
    \end{equation}
    where $d(t)$ is a time-varying minimum distance;
    \item[\textbf{O3.}] \emph{Docking:} After reaching target positions, modules physically connect via the latching mechanism, requiring close proximity ($\|\bm{p}_i - \bm{p}_j\| \approx L$ for adjacent modules).
\end{enumerate}

Objective~\textbf{O2} introduces a time-varying safety distance $d(t)$ to reconcile the competing requirements of safe navigation and close-range docking. During navigation, $d(t) = d_\text{out}$ provides a conservative safety margin. During the docking phase, $d(t)$ is reduced to $d_\text{in} < d_\text{out}$, allowing modules to approach within latching range.

%% file: Sections/methodology.tex

This section presents the proposed distributed ADMM-CBF framework for safe multi-agent shape formation and reconfiguration. The architecture overview is first described, followed by the distributed MPC formulation via ADMM, the CBF safety filter, the distance adaptation mechanism, and the task coordination strategy.
\subsection{Architecture Overview}
\label{sec:architecture}

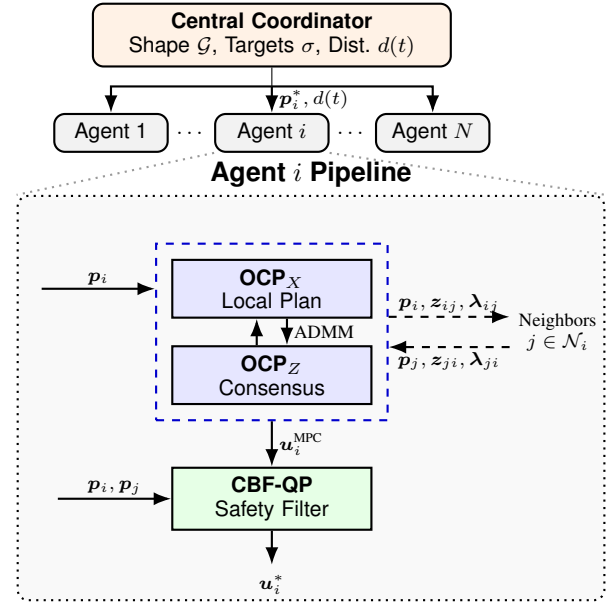
\begin{figure}[t]
    \centering
    \input{Figures/architecture_fig.tex}
    \caption{Architecture of the proposed ADMM-CBF framework. A central assignment provides target positions, while distributed MPC solved via ADMM generates coordinated trajectories through inter-agent communication. A local CBF-based safety filter ensures collision avoidance before control inputs are applied to each robot.}
    \label{fig:architecture}
\end{figure}

The proposed framework combines distributed model predictive control with a distributed safety filter, coordinated by a lightweight central planner. Fig.~\ref{fig:architecture} illustrates the system architecture. A central coordinator provides target shape commands and position assignments to all modules. Each module then independently executes the following pipeline at every control step:
\begin{enumerate}
    \item Solve a local trajectory optimization problem (OCP$_X$) to plan a path toward the assigned target;
    \item Exchange planned trajectories with neighbors and solve a consensus subproblem (OCP$_Z$);
    \item Update the ADMM dual variables;
    \item Filter the resulting velocity command through a CBF-based safety filter;
    \item Send the safe velocity reference to the low-level controller.
\end{enumerate}
The ADMM planner provides an approximate motion plan with a receding horizon, giving the CBF sufficient look-ahead information to avoid deadlocks. The CBF, in turn, compensates for the lack of convergence guarantees in the ADMM scheme by enforcing safety at every control step. This separation of concerns is the key design principle: ADMM handles planning, CBF handles safety.

\subsection{Distributed MPC via ADMM}
\label{sec:admm}

The trajectory planning problem for $N$ agents is formulated as a coupled optimal control problem (OCP), where coupling arises from inter-agent collision avoidance constraints. Following the decentralized ADMM decomposition in~\cite{ruben2017decentralised}, the problem is split into agent-local subproblems connected through consensus constraints on local copies of neighboring agents' trajectories.

\subsubsection{Trajectory Subproblem (OCP$_X$)}

Each agent $i$ solves a local trajectory optimization over a horizon $T_f$ with $N_h$ shooting intervals:
\begin{subequations}
\label{eq:ocpX}
\begin{align}
    \min_{\bm{p}_i(\cdot),\, \bm{u}_i(\cdot)} \quad & J_i^X = \sum_{k=0}^{N_h} \ell_i(\bm{p}_{i,k}) + \sum_{k=0}^{N_h-1} \|\bm{u}_{i,k}\|^2 \label{eq:ocpX_cost} \\
    & + \sum_{k=0}^{N_h} \left( \bm{\lambda}_i^{k\top} (\bm{z}_i^k - \bm{p}_{i}^k) + \tfrac{\mu}{2} \|\bm{z}_i^k - \bm{p}_{i}^k\|^2 \right) \nonumber \\
    & + \sum_{j \in \mathcal{N}_i} \sum_{k=0}^{N_h} \left( \bm{\lambda}_{ji}^{k\top} (\bm{z}_{ji}^k - \bm{p}_{i}^k) + \tfrac{\mu}{2} \|\bm{z}_{ji}^k - \bm{p}_{i}^k\|^2 \right) \nonumber \\
    \text{s.t.} \quad & \bm{p}_{i}^{k+1} = \bm{p}_{i}^{k} + \Delta t\, \bm{u}_{i}^{k}, \label{eq:ocpX_dyn} \\
    & \|\bm{u}_{i}^k\|_\infty \leq u_{\max}, \label{eq:ocpX_vel} \\
    & \bm{p}_{i}^{0} = \hat{\bm{p}}_i, \label{eq:ocpX_ic}
\end{align}
\end{subequations}
where $\ell_i(\bm{p}_{i,k}) = w_g\|\bm{p}_{i,k} - \bm{g}_{\sigma(i)}\|^2$ penalizes the squared distance to the assigned target with weight $w_g = 1 + |\mathcal{N}_i|$ to balance the tracking objective against the aggregate consensus penalty, $\bm{z}_i^k$ and $\bm{z}_{ji}^k$ are local copies from OCP$_Z$, $\bm{\lambda}_i^k$ and $\bm{\lambda}_{ji}^k$ are the corresponding dual variables, $\mu > 0$ is the ADMM penalty parameter, $\mathcal{N}_i = \mathcal{N} \setminus \{i\}$ is the neighbor set, and $\hat{\bm{p}}_i$ is the current position estimate. Note that OCP$_X$ contains no collision avoidance constraints; these are handled by OCP$_Z$.

\subsubsection{Collision Copy Subproblem (OCP$_Z$)}

Each agent $i$ solves a second subproblem to update its local copies while enforcing collision avoidance:
\begin{subequations}
\label{eq:ocpZ}
\begin{align}
    \min_{\bm{z}_i, \bm{z}_{ij}, \bm{s}_{ij}} \quad & J_i^Z = \sum_{k=0}^{N_h} \left( \bm{\lambda}_i^{k\top} (\bm{z}_i^k - \bm{p}_{i}^k) + \tfrac{\mu}{2}\|\bm{z}_i^k - \bm{p}_{i}^k\|^2 \right) \label{eq:ocpZ_cost} \\
    & + \sum_{j \in \mathcal{N}_i} \sum_{k=0}^{N_h} \left( \bm{\lambda}_{ij}^{k\top} (\bm{z}_{ij}^k - \bm{p}_{j}^k) + \tfrac{\mu}{2}\|\bm{z}_{ij}^k - \bm{p}_{j}^k\|^2 \right) \nonumber \\
    & + \sum_{j \in \mathcal{N}_i} \sum_{k=0}^{N_h} (s_{ij}^k)^2 \nonumber \\
    \text{s.t.} \quad & \|\bm{z}_i^k - \bm{z}_{ij}^k\|^2 + s_{ij}^k \geq d^2, \quad \forall j \in \mathcal{N}_i, \label{eq:ocpZ_coll}\\
    & s_{ij}^k \geq 0, \quad \forall j \in \mathcal{N}_i, \label{eq:ocpZ_slack}
\end{align}
\end{subequations}
where $\bm{z}_{ij}^k$ is agent~$i$'s local copy of agent~$j$'s position at step $k$, $s_{ij}^k$ is a slack variable, and $d$ is the active safety distance. The slack variable guarantees that OCP$_Z$ remains feasible even when the current copies are in close proximity, and it is penalized heavily so that it activates only when a strictly feasible collision-free copy cannot be found.

\subsubsection{Dual Variable Update}

After solving OCP$_X$ and OCP$_Z$, the dual variables are updated via the standard ADMM rule:
\begin{equation}
\label{eq:dual_update}
\begin{aligned}
    \bm{\lambda}_i^{k} &\leftarrow \bm{\lambda}_i^{k} + \mu\,(\bm{z}_i^k - \bm{p}_i^k), \\
    \bm{\lambda}_{ij}^{k} &\leftarrow \bm{\lambda}_{ij}^{k} + \mu\,(\bm{z}_{ij}^k - \bm{p}_j^k), \quad \forall j \in \mathcal{N}_i.
\end{aligned}
\end{equation}

\subsubsection{Execution Strategy}

Before the ADMM output is used for control, an initialization phase is
required. During this phase, each module maintains its position via
low-level station-keeping while the ADMM subproblems iterate in the
background. This allows the dual variables~$\bm{\lambda}$ and local
copies~$\bm{z}$ to accumulate meaningful values. Once the primal residual has
stabilized, the ADMM velocity output is passed to the CBF filter. After each OCP$_X$ solve, agent~$i$ broadcasts its planned trajectory $\bm{p}_i^{0:N_h}$ to all neighbors. After each OCP$_Z$ solve, agent~$i$ broadcasts its local copies $\bm{z}_{ij}^{0:N_h}$ and updated multipliers $\bm{\lambda}_{ij}^{0:N_h}$. This single-iteration scheme corresponds to $m = 1$ ADMM iteration per control step, which, although far from convergence, can provide a sufficient approximate plan for the CBF to operate on.

\begin{assumption} \label{assump:full_graph}
The communication and interaction graph is fully connected, such that the neighbor set for each agent $i$ is defined as $\mathcal{N}_i = \mathcal{N} \setminus \{i\}$. This topology satisfies the bidirectional communication condition required to ensure the symmetry and mathematical validity of the distributed ADMM consensus updates \cite{ruben2017decentralised}.
\end{assumption}


\subsection{Distributed CBF Safety Filter}
\label{sec:cbf_filter}

Because ADMM convergence is only guaranteed for convex problems~\cite{boyd2011admm}, and the collision avoidance constraints in OCP$_Z$ are nonconvex, there is no formal guarantee that the resulting ADMM trajectories are collision-free. To recover a rigorous safety guarantee, each agent applies a discrete safety filter to the ADMM output. At each control step, let $\bm{u}_i^\text{MPC}$ denote the velocity command from OCP$_X$. To satisfy the decoupled safety constraint derived in Section~\ref{sec:cbf_theory}, agent $i$ computes its actual safe velocity $\bm{u}_i^*$ by solving the following quadratic program (QP):
\begin{subequations}
\label{eq:cbf_qp}
\begin{align}
    \min_{\bm{u}_i} \quad & \|\bm{u}_i - \bm{u}_i^\text{MPC}\|^2 \label{eq:cbf_qp_cost} \\
    \text{s.t.} \quad & 2(\bm{p}_i - \bm{p}_j)^\top \bm{u}_i + \beta\gamma\, h_{ij}(t) \geq 0, \quad \forall j \in \mathcal{N}_i, \label{eq:cbf_qp_cbf} \\
    & \|\bm{u}_i\|_\infty \leq u_{\max}, \label{eq:cbf_qp_vel}
\end{align}
\end{subequations}
where $h_{ij}(t) = \|\bm{p}_i - \bm{p}_j\|^2 - d(t)^2$. By using a time-varying distance $d(t)$, the filter dynamically adapts between the conservative safety margins needed for free-space navigation and the close-proximity requirements for docking. Because the safety constraints are decoupled and linear with respect to $\bm{u}_i$, the QP~\eqref{eq:cbf_qp} is strictly convex and solvable in microseconds. By minimizing the deviation from the ADMM reference, the QP acts as a minimally invasive safety filter. The resulting safe velocity $\bm{u}_i^*$ is then passed directly to the low-level controller for execution.

\subsection{Distance Adaptation for Docking}
\label{sec:distance}

Self-reconfiguration requires modules to both navigate safely and dock at close range. These objectives impose conflicting requirements on the safety distance $d(t)$: a large $d$ is desirable for safe transit, but modules must approach within the latching range ($\approx L$) to physically connect. To resolve this conflict, the safety distance is adapted in two phases:
\begin{equation}
\label{eq:distance}
    d(t) = \begin{cases}
        d_\text{out}, & \text{navigation phase,} \\
        d_\text{in},  & \text{docking phase,}
    \end{cases}
\end{equation}
where $d_\text{out} > d_\text{in} > 0$. The transition is triggered by the central coordinator when all modules are within a tolerance $\epsilon$ to their assigned positions. Each module signals its arrival, and the coordinator responds with a single broadcast to update $d(t)$, keeping the central intervention sparse. Specifically, $d_\text{out}$ is used during the initial approach, defined as the module diagonal plus a safety margin, while $d_\text{in}$ can be set at the module side length $L$, or slightly below, to permit physical contact on the module faces. Both OCP$_Z$~\eqref{eq:ocpZ_coll} and the CBF~\eqref{eq:cbf_qp_cbf} are updated with the active $d(t)$ at each step, ensuring safety is maintained under both regimes.

\subsection{Task Coordination}
\label{sec:coordination}

The overall reconfiguration task is orchestrated by a central coordinator that performs only sparse, high-level operations:
\begin{enumerate}
    \item Transmit the target shape $\mathcal{G}$ to all modules;
    \item Compute the position assignment $\sigma$ via the Hungarian algorithm, minimizing total displacement $\sum_{i} \|\hat{\bm{p}}_i - \bm{g}_{\sigma(i)}\|^2$;
    \item Broadcast the active distance $d(t)$ and latching commands.
\end{enumerate}
All real-time computation (trajectory planning, collision avoidance, and safety filtering) is performed onboard each module. For self-reconfiguration across multiple shapes $\mathcal{G}^{(1)} \to \mathcal{G}^{(2)} \to \cdots$, the coordinator issues delatch commands, waits for modules to separate, updates the target shape and assignment, and repeats the process. Algorithm~\ref{alg:admm_cbf} summarizes the complete framework.

\begin{algorithm}[h]
\caption{Distributed ADMM-CBF Framework}
\label{alg:admm_cbf}
\begin{algorithmic}[1]
\REQUIRE Target shape $\mathcal{G}$, assignment $\sigma$, distance $d(t)$
\STATE Initialize $\bm{\lambda}_i^k, \bm{\lambda}_{ij}^k \leftarrow \bm{0}$, $\bm{z}_i^k, \bm{z}_{ij}^k \leftarrow \hat{\bm{p}}_i$
\WHILE{modules not at target positions}
    \STATE \textbf{Each agent $i$ in parallel:}
    \STATE Solve OCP$_X$~\eqref{eq:ocpX} $\to$ trajectory $\bm{p}_i^{0:N_h}$
    \STATE Broadcast $\bm{p}_i^{0:N_h}$ to neighbors
    \STATE Receive neighbors' trajectories $\bm{p}_j^{0:N_h}$
    \STATE Solve OCP$_Z$~\eqref{eq:ocpZ} $\to$ copies $\bm{z}_i^{0:N_h}$, $\bm{z}_{ij}^{0:N_h}$
    \STATE Update duals~\eqref{eq:dual_update}
    \STATE Set $\bm{u}_i^\text{MPC} \leftarrow \bm{u}_i^0$ from OCP$_X$
    \STATE Solve CBF-QP~\eqref{eq:cbf_qp} with $d(t)$ $\to$ $\bm{u}_i^*$
    \STATE Apply $\bm{u}_i^*$ to low-level controller
\ENDWHILE
\STATE Activate latching mechanism
\end{algorithmic}
\end{algorithm}

%% file: Figures/architecture_fig.tex
\begin{tikzpicture}[
    >=Latex,
    font=\sffamily\footnotesize,
    central/.style={rectangle, draw=black, thick, fill=orange!10, text width=4.5cm, align=center, minimum height=0.7cm, rounded corners},
    agentbox/.style={rectangle, draw=black, thick, fill=gray!10, minimum width=1.5cm, minimum height=0.5cm, rounded corners},
    algoblock/.style={rectangle, draw=black, thick, fill=blue!10, text width=2.4cm, align=center, minimum height=0.7cm},
    cbfblock/.style={rectangle, draw=black, thick, fill=green!10, text width=2.4cm, align=center, minimum height=0.7cm},
    flowline/.style={thick, ->},
    commline/.style={thick, dashed, ->}
]

\node[central] (coordinator) {\textbf{Central Coordinator} \\ Shape $\mathcal{G}$, Targets $\sigma$, Dist. $d(t)$};

\node[agentbox, below=0.6cm of coordinator] (agenti) {Agent $i$};
\node[agentbox, left=0.6cm of agenti] (agent1) {Agent 1};
\node[agentbox, right=0.6cm of agenti] (agentN) {Agent $N$};

\node at ($(agent1)!0.5!(agenti)$) {$\dots$};
\node at ($(agenti)!0.5!(agentN)$) {$\dots$};

\draw[-] (coordinator.south) -- ++(0,-0.25) coordinate (branch);
\draw[flowline] (branch) -| (agent1.north);
\draw[flowline] (branch) -| (agentN.north);
\draw[flowline] (branch) -- (agenti.north) node[midway, right=-1pt, font=\scriptsize] {$\bm{p}_i^*, d(t)$};


\node[algoblock, below=1.4cm of agenti] (ocpx) {\textbf{OCP}$_X$ \\ Local Plan};
\node[algoblock, below=0.35cm of ocpx] (ocpz) {\textbf{OCP}$_Z$ \\ Consensus};

\draw[flowline] ($(ocpx.south)+(0.2,0)$) -- ($(ocpz.north)+(0.2,0)$) node[midway, right=-1pt, font=\scriptsize] {ADMM};
\draw[flowline] ($(ocpz.north)+(-0.2,0)$) -- ($(ocpx.south)+(-0.2,0)$);

\node[draw=blue!80!black, thick, dashed, inner sep=0.2cm, fit=(ocpx) (ocpz)] (admm) {};

\node[cbfblock, below=0.6cm of admm] (cbf) {\textbf{CBF-QP} \\ Safety Filter};

\draw[flowline] (admm.south) -- (cbf.north) node[midway, right=-1pt, font=\scriptsize] {$\bm{u}_i^\text{MPC}$};

\draw[flowline] (cbf.south) -- ++(0,-0.45) node[below, font=\scriptsize] {$\bm{u}_i^*$};

\draw[flowline] ([xshift=-1.5cm]admm.west |- ocpx.west) -- (admm.west |- ocpx.west) node[midway, above=-2pt, font=\scriptsize] {$\bm{p}_i$};
\draw[flowline] ([xshift=-1.5cm]cbf.west) -- (cbf.west) node[midway, above=-2pt, font=\scriptsize] {$\bm{p}_i, \bm{p}_j$};

\node[right=1.6cm of admm, align=center, font=\scriptsize] (neigh) {Neighbors \\ $j \in \mathcal{N}_i$};

\draw[commline] ([yshift=0.2cm]admm.east) -- ([yshift=0.2cm]neigh.west) node[midway, above=-2pt, font=\scriptsize] {$\bm{p}_i, \bm{z}_{ij}, \bm{\lambda}_{ij}$};
\draw[commline] ([yshift=-0.2cm]neigh.west) -- ([yshift=-0.2cm]admm.east) node[midway, below=-2pt, font=\scriptsize] {$\bm{p}_j, \bm{z}_{ji}, \bm{\lambda}_{ji}$};

\begin{scope}[on background layer]
    \coordinate (boxnw) at ([xshift=-1.7cm, yshift=0.5cm]admm.north west);
    \coordinate (boxse) at ([xshift=0.5cm, yshift=-0.8cm]neigh.south |- cbf.south);
    
    \node[draw=black, thick, dotted, fill=gray!5, rounded corners, fit=(boxnw) (boxse)] (agentibox) {};
    \node[font=\sffamily\bfseries, anchor=south, inner sep=3pt] at (agentibox.north) {Agent $i$ Pipeline};
\end{scope}

\draw[thick, dotted, color=gray!80] (agenti.south west) -- (agentibox.north west);
\draw[thick, dotted, color=gray!80] (agenti.south east) -- (agentibox.north east);

\end{tikzpicture}

%% file: Sections/scalability.tex
\label{sec:scalability}

To rigorously evaluate the proposed framework, we conducted a series of simulated self-assembly scenarios. The primary objectives are to demonstrate the necessity of both the predictive (ADMM) and reactive (CBF) components, and verify the computational scalability of the distributed formulation. 

\subsection{Simulation Setup}
The simulations mirror the physical constraints of the robotic modules. The framework is implemented in Python, utilizing \texttt{CasADi} \cite{andersson2019casadi}, the \texttt{Fatrop} \cite{vanroye2023fatrop} solver for the ADMM optimal control problems, and \texttt{OSQP}~\cite{osqp} for the CBF quadratic program. The system operates at a control frequency of $5$ Hz ($dt = 0.2$ s), establishing a target real-time computational budget of $200$ ms per step. Across all scenarios, the module actuator saturation is constrained to $u_{max} = 0.04$ m/s. The core controller parameters are kept constant across all experiments: a predictive horizon of $N_{h} = 10$, an ADMM penalty parameter of $\mu = 10$, and CBF parameters of $\gamma = 1$ and $\beta = 0.5$. The experiments evaluate swarm sizes $N \in \{4, 8, 9, 16, 25\}$. The proposed ADMM-CBF method is compared against two baselines:
\begin{enumerate}
    \item ADMM-only \cite{ruben2017decentralised}: A decentralized MPC without the continuous-time safety filter, representing purely predictive planning.
    \item CBF-only \cite{multiCBF}: A strictly reactive approach where modules navigate towards goals while relying exclusively on the CBF quadratic program for collision avoidance.
\end{enumerate}

For the ADMM-based methods, an initialization phase of $50$ steps is executed before the modules begin moving to establish an initial consensus. All reported computational times are measured strictly during the online phase (post-warm-up) to accurately reflect real-time operational performance. To eliminate initial-condition bias, the scenarios are evaluated across multiple randomized initial spatial distributions (seeds). Specifically, the ablation study uses $5$ seeds per configuration, and the scalability analysis uses $3$ seeds. All reported metrics represent the aggregate performance across these runs. An additional stress test with $N=64$ agents is presented in Section~\ref{sec:scalability_limits} to explore the computational boundaries of the framework.

\subsection{Ablation Study}
To evaluate the robustness of the framework during sustained operations, the ablation study focuses on complex, sequential multi-shape reconfigurations. The swarm is tasked to transition through three distinct lattice shapes for swarm sizes $N \le 16$, and two distinct shapes for the $N=25$ case. Table \ref{tab:ablation_results} summarizes the success, safety (collision-free), and deadlock rates. The results highlight a fundamental trade-off between the baselines.

    \subsubsection{Safety violations in ADMM-only} Even at smaller swarm sizes ($N \le 9$), the purely predictive ADMM experiences collisions. Because the multi-agent collision avoidance problem is highly nonconvex, and the distributed algorithm is restricted to a single iteration per control step to meet real-time constraints, the ADMM frequently fails to converge to a safe consensus. This lack of exact convergence yields planned trajectories that violate physical boundaries.
    \subsubsection{Deadlocks in CBF-only} The reactive CBF filter alone achieves safe navigation and successful assembly for smaller swarms ($N \le 9$), where the density is low enough such that the modules are not frequently obstructed. Its limitation is foresight: while it guarantees $100\%$ safety, it is highly susceptible to deadlocks as swarm density increases. At $N=16$ and $N=25$, the modules reached configurations where they cannot navigate past each other, resulting in deadlocks.
    \subsubsection{ADMM-CBF framework} The proposed framework leverages the complementary strengths of both approaches to resolve these limitations. While ADMM lacks formal safety guarantees, it provides an approximate plan. The CBF then filters this approximate reference, enforcing safety at each control step. This synergy results in a $100\%$ success and safety rate across all tested configurations. The ADMM plan carries the multi-agent coordination that a reactive filter lacks, while the CBF provides the safety guarantee that the approximate plan cannot. Thus, the proposed combination remains both safe and deadlock-free across all evaluated swarm sizes.

\begin{table}[tb]
\centering
\caption{Ablation Study: Success, Safety, and Deadlock Rates}
\label{tab:ablation_results}
\resizebox{\columnwidth}{!}{%
\begin{tabular}{@{} l rrr rrr rrr @{}}
\toprule
& \multicolumn{3}{c}{\textbf{ADMM}} & \multicolumn{3}{c}{\textbf{CBF}} & \multicolumn{3}{c}{\textbf{ADMM-CBF}} \\
\cmidrule(lr){2-4} \cmidrule(lr){5-7} \cmidrule(lr){8-10}
$N$ & Succ. & Safe & Dead. & Succ. & Safe & Dead. & Succ. & Safe & Dead. \\
\midrule
4  & 100\% & 0\% & 0\% & 100\% & 100\% &   0\% & \textbf{100\%} & \textbf{100\%} & \textbf{0\%} \\
8  & 100\% & 0\% & 0\% & 100\% & 100\% &   0\% & \textbf{100\%} & \textbf{100\%} & \textbf{0\%} \\
9  & 100\% & 0\% & 0\% & 100\% & 100\% &   0\% & \textbf{100\%} & \textbf{100\%} & \textbf{0\%} \\
16 & 100\% & 0\% & 0\% &  20\% & 100\% &  80\% & \textbf{100\%} & \textbf{100\%} & \textbf{0\%} \\
25 &  100\%   & 0\%  & 0\%  &   0\% & 100\% & 100\% & \textbf{100\%} & \textbf{100\%} & \textbf{0\%} \\
\bottomrule
\multicolumn{10}{l}{\scriptsize \textit{Succ.} = Success Rate, \textit{Safe} = Collision-Free Rate, \textit{Dead.} = Deadlock Rate.}
\end{tabular}%
}
\end{table}

\subsection{Computational Scalability and Real-Time Feasibility}
\begin{figure}[tb]
    \centering
    \includegraphics[width=0.95\linewidth]{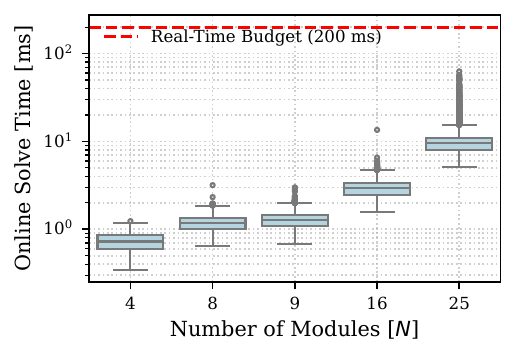}
    \caption{Per-agent online computation time of the proposed ADMM-CBF framework across varying swarm sizes. Each box shows the distribution of $t_{\text{total}} = t_{\text{OCP}_X} + t_{\text{OCP}_Z} + t_{\text{CBF}}$ across all agents and post-warmup time steps.}
    \label{fig:scalability}
\end{figure}

To evaluate the computational performance of the framework, we recorded the execution times of single-shape self-assembly scenarios across the same set of swarm sizes ($N \in \{4, 8, 9, 16, 25\}$). All simulations were executed on a laptop equipped with an Intel Core Ultra 7 265H processor and 64 GB of RAM. Figure \ref{fig:scalability} illustrates the distribution of the per-agent online solve times ($t_{\text{total}} = t_{\text{OCP}_X} + t_{\text{OCP}_Z} + t_{\text{CBF}}$).

The framework satisfies the $200$ ms real-time budget for all evaluated swarm sizes. For $N=4$, the mean total per-agent solve time per step is approximately $0.75$ ms. As the swarm scales to $N=25$, the mean per-agent solve time increases to $9.55$ ms, with a maximum of 62.66 ms. The majority of the computation time comes from the dual consensus update (OCP$_Z$). As the spatial density of the swarm increases, the number of neighbors within the interaction radius grows. This directly increases the number of collision avoidance constraints that the Nonlinear Programming (NLP) solver must evaluate per module, causing the OCP$_Z$ mean per-agent solve times to rise nonlinearly from $0.39$ ms at $N=4$ to $8.93$ ms at $N=25$. The mean per-agent solve time for the trajectory subproblem (OCP$_X$) also grows marginally as $N$ increases, rising from $0.35$ ms to $0.61$ ms. This is due to the increasing number of terms in the objective function as the local neighborhood expands. Conversely, the CBF reactive filter remains computationally light, requiring less than $0.02$ ms across all evaluated sizes. While the reliance on a full interaction neighborhood introduces theoretical scaling limits, results confirm that the maximum per-agent per-step computation time for configurations up to $N=25$ remained strictly within the $200$ ms budget. These results confirm that with dedicated solvers, the ADMM-CBF formulation can comfortably satisfy real-time constraints, even on standard computing hardware.

\subsection{Scalability Limitations} \label{sec:scalability_limits}
While the framework successfully resolves collisions and deadlocks for swarms up to $N=25$, testing with larger agent counts reveals specific algorithmic limitations.

\subsubsection{Horizon Length vs. Computation Time} In scenarios with $N=64$ agents, short predictive horizons ($N_{h}=10$, $t_f=2.0$ s) are insufficient to eliminate all deadlocks. The short prediction fails to guide agents out of complex local minima. A subsequent simulation suggests that extending the horizon ($N_{h}=25$,$t_f=5.0$ s), and increasing the ADMM consensus penalty ($\mu=200$), can resolve challenging deadlocks and complete the assembly. However, this extension significantly increases the computational burden of the OCP$_Z$ step. The mean OCP$_Z$ solve time grew from $115$ ms in the short-horizon case to $310$ ms in the long-horizon case, with occasional peaks exceeding $3$ seconds. This renders the extended formulation impractical for real-time implementation. A potential solution is an asynchronous architecture, where the ADMM planner operates at a lower frequency ($1-2$ Hz) to compute longer horizon plans, while the CBF filter operates at a higher frequency ($\ge 50$ Hz) to guarantee safety.

\subsubsection{Neighborhood Size Limits} The current implementation relies on a full interaction neighborhood. As more agents cluster together during assembly, the number of coupling constraints evaluated by the NLP solver grows rapidly. Dynamically defining and managing limited communication neighborhoods within ADMM-based distributed MPC remains an open challenge~\cite{ruben2017decentralised,Tian2022}, and addressing it is essential for scaling beyond the swarm sizes evaluated in this work.

\begin{figure*}[t]
    \centering
    \includegraphics[width=\linewidth]{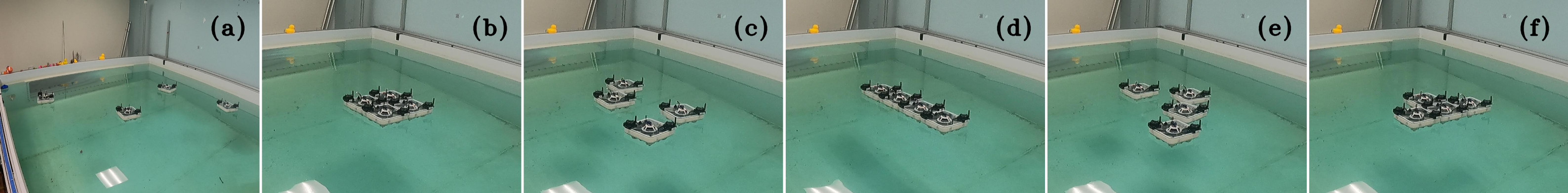}
    \caption{Sequential snapshots of the four-module self-reconfiguration experiment in the water tank. Left to right: (a) The scattered initial configuration at $t = 40$\,s. (b) The modules successfully dock into the initial $2 \times 2$ square lattice. (c) The swarm unlatches and actively negotiates collision-free trajectories to reconfigure. (d) The fully assembled $4 \times 1$ line. (e) The modules break the line formation and navigate toward their final assignment. (f) The final latched 3-1 T-shape.}
    \label{fig:experiment}
\end{figure*}

\subsubsection{Centralized Target Assignment} The framework currently computes target assignments centrally before motion begins. In dense swarms, intersecting trajectories inevitably create physical bottlenecks. Resolving these deadlocks locally could involve integrating task assignment directly into the control layer. Implementing distributed techniques, such as local task swapping~\cite{TaskSwapping}, would allow blocked agents to dynamically exchange targets. By negotiating targets online, the system could escape deadlocks without relying on extended prediction horizons. Ultimately, this research direction could eliminate the need for receding horizon planning entirely, paving the way for highly scalable, purely reactive, safe CBF-based architectures.

%% file: Sections/experiments.tex
\label{sec:experiments}

\begin{figure}[t]
    \centering
    \includegraphics[width=0.95\linewidth]{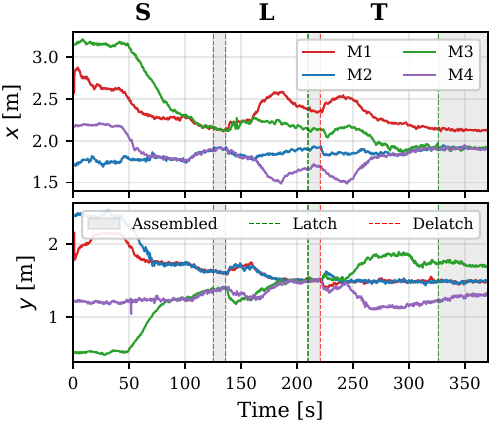}
    \caption{Module positions during a three-shape self-reconfiguration experiment with four modules. The modules form a square lattice~(S), reconfigure into a line~(L), and reconfigure into a T-shape~(T). Gray regions indicate assembled (latched) periods. Green and red dashed lines mark latch and delatch events, respectively.}
    \label{fig:trajectories}
\end{figure}

We validate the proposed ADMM-CBF framework on four physical modules in an indoor water tank ($4.0 \times 2.7 \times 1.2$\,m) using ultrasonic beacon localization.  Each module runs the ADMM planner and CBF filter onboard a Raspberry Pi~4 using \texttt{CasADi} and \texttt{IPOPT}~\cite{wachter2006ipopt}, targeting a control frequency of 5 Hz. The ADMM uses a horizon of $T_f = 2$\,s with $N_h = 10$ shooting intervals, penalty $\mu = 10$, and one ADMM iteration per control step ($m=1$). The CBF gain is $\gamma = 1$, with $\beta = 0.5$, and the velocity bound is $u_\text{max} = 0.04$\,m/s. The safety distances are $d_\text{out} = 0.32$\,m and $d_\text{in} = 0.18$\,m. Localization and low-level control runs at a higher frequency of 50 Hz. The task is to perform a three-shape self-reconfiguration sequence depicted in  Fig.~\ref{fig:experiment} and Fig.~\ref{fig:trajectories}: a $2 \times 2$ square lattice~(S), a $4 \times 1$  line~(L), and a 3-1 T-shape~(T).

Table~\ref{tab:timing} summarizes the computation times. The ADMM cycle  (OCP$_X$ + OCP$_Z$) has a median computation time of $149.7$\,ms, fitting  within the $200$\,ms budget for $86\%$ of time steps. The remaining $14\%$  exceeds the budget due to occasional solver delays, predominantly from one  module with consistently higher solve times. The CBF filter adds a median  overhead of $32.4$\,ms. We note that the use of \texttt{IPOPT}, a general-purpose interior-point solver, leaves significant room for improvement through dedicated solvers such as \texttt{Fatrop} and \texttt{OSQP}, as shown in the simulation experiments.

\begin{figure}[t]
    \centering
    \includegraphics[width=0.95\linewidth]{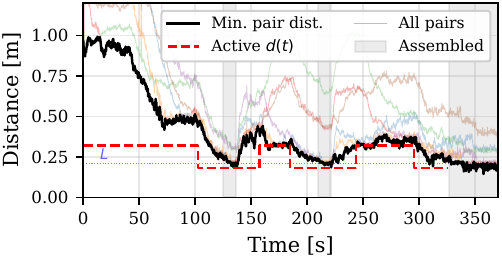}
    \caption{Inter-agent distances during the experiment. The black line shows the minimum pairwise distance across all module pairs at each time step; colored lines show individual pairs. The red dashed line indicates the active safety distance~$d(t)$, which switches between $d_\text{out} = 0.32$\,m (navigation) and $d_\text{in} = 0.18$\,m (docking approach). During assembled periods (gray), modules are physically latched and distances are not safety-relevant. The blue dotted line marks the module side length $L = 0.21$\,m.}
    \label{fig:distances}
\end{figure}

Fig.~\ref{fig:distances} shows the inter-agent distances throughout the  experiment. During navigation ($d(t) = d_\text{out} = 0.32$\,m) and docking phases ($d(t) = d_\text{in} = 0.18$\,m), the  minimum measured distance across all module pairs is maintained above the active safety threshold, suggesting that the ADMM-CBF approach is capable of providing safety measures. However, some instances fall below the safety distance $d(t)$, and at times even below the physical module side $L$, attributable mainly to the localization uncertainty of the acoustic beacon system (estimated at $\pm 2$--$3$\,cm), as well as model uncertainty and low-level control error. Robust CBF formulations \cite{9636584} can be applied to account for such bounded uncertainty.

\begin{table}[t]
    \centering
    \caption{Computation times on embedded hardware.
    }
    \label{tab:timing}
    \setlength{\tabcolsep}{5pt}
    \begin{tabular}{l c c c c}
        \toprule
        \textbf{Solver} & \textbf{Median} & \textbf{Mean} & \textbf{P95} & \textbf{Max} \\
                        & {[ms]}          & {[ms]}        & {[ms]}       & {[ms]} \\
        \midrule
        OCP$_X$                 &  40.9 &  44.8 &  76.1 & 348.1 \\
        OCP$_Z$                 & 106.8 & 115.8 & 184.0 & 496.7 \\
        \midrule
        ADMM cycle$^\dagger$    & 149.7 & 160.6 & 247.2 & 639.5 \\
        \midrule
        CBF filter              &  32.4 &  35.4 &  63.5 & 279.4 \\
        \bottomrule
        \multicolumn{5}{l}{\footnotesize $^\dagger$\,Paired OCP$_X$ + OCP$_Z$. Within 200\,ms budget for 86\% of steps.}
    \end{tabular}
\end{table}

